\documentclass[conference]{IEEEtran}
\usepackage[shortlabels]{enumitem}
\usepackage[nospace,noadjust]{cite}
\usepackage{graphicx} 
\usepackage{epsfig} 
\usepackage{amsmath,amsthm,amssymb}  
\usepackage{float}
\usepackage{cases,setspace,adjustbox}
\usepackage{array}
\usepackage{bbm}
\usepackage{booktabs}
\usepackage[update,prepend]{epstopdf}
\usepackage{eqparbox}
\usepackage{url}
\usepackage{caption, subcaption}
\usepackage{xcolor}
\usepackage{mathtools}
\usepackage{bbm}
\usepackage{multirow}
\usepackage{multicol}

\usepackage[acronym]{glossaries}
\usepackage{comment}
\usepackage{xcolor}
\newcommand*\circled[1]{\kern-2.5em%
  \put(0,4){\color{white}\circle*{11}}\put(0,4){\circle{11}}%
  \put(-3,0){\color{black}\large#1}~~}
\setacronymstyle{long-short}

\newacronym{5G}{5G}{fifth-generation}
\newacronym{6G}{6G}{sixth-generation}
\newacronym{c2}{C2}{command-and-control}
\newacronym{1dcnn}{1D-CNN}{one-dimensional Convolutional Neural Network}
\newacronym{AR}{AR}{Augmented Reality}
\newacronym{ai}{AI}{Artificial Intelligence}
\newacronym{arima}{ARIMA}{Autoregressive Integrated Moving Average}
\newacronym{ae}{AE}{autoencoder}

\newacronym{cdf}{CDF}{cumulative distribution function}
\newacronym{cnn}{CNN}{Convolutional Neural Network}
\newacronym{comp}{CoMP}{Coordinated Multi-Point}
\newacronym{convlstm}{ConvLSTM}{Convolutional LSTM}
\newacronym{cp}{CP}{Cyclic Prefix}
\newacronym{cpd}{CPD}{Change Point Detection}
\newacronym{crs}{CRS}{Cell Specific
Reference Signal}
\newacronym{ctl}{CTL}{Communications Technology Laboratory}
\newacronym{CU}{CU}{Centralized Unit}
\newacronym{cusum}{CUSUM}{Cumulative Sum Control Chart}
\newacronym{cv}{CV}{Cross-Validation}

\newacronym{d2d}{D2D}{Device-to-Device}
\newacronym{dc}{DC}{dual connectivity}
\newacronym{dci}{DCI}{Downlink Control Information}
\newacronym{DL}{DL}{Deep Learning}
\newacronym{dl}{DL}{downlink}
\newacronym{dsrc}{DSRC}{Dedicated short-range communications}
\newacronym{dss}{DSS}{Dynamic Spectrum Sharing}
\newacronym{DU}{DU}{Distributed Unit}

\newacronym{ets}{ETS}{Exponential Smoothing}
\newacronym{embb}{eMBB}{enhanced mobile broadband}
\newacronym{enb}{eNB}{Evolved Node-B}
\newacronym{epc}{EPC}{Evolved Packet Core}
\newacronym{emtc}{eMTC}{enhanced MTC}

\newacronym{fdd}{FDD}{Frequency Domain Duplex}
\newacronym{fdm}{FDM}{Frequency Domain Multiplexing}
\newacronym{firstnet}{FirstNet}{First Responder Network Authority}

\newacronym{gan}{GAN}{Generative Adversarial Network}
\newacronym{gps}{GPS}{Global Positioning System}
\newacronym{gnb}{gNB}{Next Generation Node-B}
\newacronym{gbps}{Gb/s}{gigabits per second}
\newacronym{gru}{GRU}{Gated Recurrent Unit}

\newacronym{harq}{HARQ}{Hybrid Automatic Repeat reQuest}
\newacronym{haps}{HAPS}{High-Altitude Platform Stations}

\newacronym{icmp}{ICMP}{Internet Control Message Protocol}
\newacronym{iot}{IoT}{Internet of Things}
\newacronym{iiot}{IIoT}{Industrial Internet of Things}
\newacronym{iomt}{IoMT}{Internet of Medical Things}
\newacronym{irs}{IRS}{Intelligent Reflecting Surfaces}
\newacronym{its}{ITS}{Intelligent Transport Systems}
\newacronym{itu}{ITU}{International Telecom Union}
\newacronym{iq}{IQ}{In-Phase and Quadrature}

\newacronym{kpi}{KPI}{Key Performance Indicator}
\newacronym{kpis}{KPIs}{Key Performance Indicators}
\newacronym{kl}{KL}{Kullback-Leibler}

\newacronym{loo}{LOO}{leave-one-out}
\newacronym{lte}{LTE}{Long Term Evolution}
\newacronym{lteapro}{LTE-A~Pro}{LTE Advanced Pro}
\newacronym{lstm}{LSTM}{Long Short-Term Memory}
\newacronym{lr}{LR}{Logistic Regression}

\newacronym{mad}{MAD}{Median Absolute Deviation}
\newacronym{mcs}{MCS}{Modulation and Coding Scheme}
\newacronym{mcptt}{MCPTT}{Mission-Critical Push-to-Talk}
\newacronym{mmimo}{mMIMO}{massive Multiple-Input Multiple-Output}
\newacronym{mmtc}{mMTC}{massive machine-type communication}
\newacronym{mtc}{MTC}{Machine-Type Communication}
\newacronym{mlp}{MLP}{Multilayer Perceptron}
\newacronym{ma}{MA}{Moving Average}
\newacronym{mm}{MM}{Moving Median}
\newacronym{ml}{ML}{Machine Learning}
\newacronym{mse}{MSE}{mean squared error}

\newacronym{nist}{NIST}{National Institute of Standards and Technology}
\newacronym{nbiot}{NB-IoT}{Narrowband-Internet of Things}
\newacronym{nr}{NR}{New Radio}
\newacronym{ntia}{NTIA}{National Telecommunications and Information Administration}
\newacronym{ntn}{NTN}{Non-Terrestrial Networks}

\newacronym{oran}{Open RAN}{Open Radio Access Network}
\newacronym{owl}{OWL}{Online Watcher of LTE}

\newacronym{pelt}{PELT}{Pruned Exact Linear Time}
\newacronym{prosas}{ProSAS}{Proactive Spectrum Adaptation Scheme}
\newacronym{prb}{PRB}{Physical Resource Block}
\newacronym{psc}{PSC}{Public Safety Communications}
\newacronym{pscch}{PSCCH}{Physical Sidelink Control Channel}
\newacronym{psfch}{PSFCH}{Physical Sidelink Feedback Channel}
\newacronym{pssch}{PSSCH}{Physical Sidelink Shared Channel}
\newacronym{pr}{PR}{precision-recall}

\newacronym{qos}{QoS}{Quality of Service}

\newacronym{ran}{RAN}{Radio Access Network}
\newacronym{rapp}{rApp}{RAN Application}
\newacronym{rans}{RANs}{Radio Access Networks}
\newacronym{rat}{RAT}{Radio Access Technology}
\newacronym{re}{RE}{Resource Element}
\newacronym{rf}{RF}{Radio Frequency}
\newacronym{ric}{RIC}{RAN Intelligent Controller}
\newacronym{ris}{RIS}{Reconfigurable Intelligent Surfaces} 
\newacronym{relu}{ReLU}{Rectified Linear Unit}
\newacronym{rl}{RL}{Reinforcement Learning}
\newacronym{rmse}{RMSE}{root-mean-square error}
\newacronym{rnn}{RNN}{Recurrent Neural Network}
\newacronym{rssi}{RSSI}{Received Signal Strength Indicator}
\newacronym{rsrp}{RSRP}{Reference Signal Received Power}
\newacronym{RU}{RU}{Radio Unit}
\newacronym{rt}{RT}{Real-Time}
\newacronym{roc}{ROC}{Receiver Operating Characteristic}

\newacronym{sdr}{SDR}{Software Defined Radio}
\newacronym{sem}{SEM}{Statistical Ensemble Method}
\newacronym{sl}{SL}{sidelink}
\newacronym{smo}{SMO}{Service and Management Orchestration}
\newacronym{snr}{SNR}{Signal-to-Noise Ratio}
\newacronym{ssb}{SSB}{Synchronization Signal Block}
\newacronym{svm}{SVM}{Support Vector Machine}

\newacronym{tcn}{TCN}{Temporal Convolutional Network}
\newacronym{td3}{TD3}{Twin Delayed Deep Deterministic Policy Gradient}
\newacronym{tdd}{TDD}{Time Domain Duplex}
\newacronym{tdm}{TDM}{Time Domain Multiplexing}
\newacronym{timegan}{TimeGAN}{Time-Series Generative Adversarial Network}
\newacronym{tr}{TR}{Technical Report}
\newacronym{thz}{THz}{terahertz}
\newacronym{TIP}{TIP}{Telecom Infra Project}

\newacronym{urllc}{URLLC}{ultra-reliable low-latency communication}
\newacronym{uav}{UAVs}{Unmanned Aerial Vehicles}
\newacronym{ue}{UE}{User Equipment}
\newacronym{ul}{UL}{uplink}

\newacronym{v2v}{V2V}{Vehicle-to-Vehicle}
\newacronym{v2x}{V2X}{Vehicle-to-Everything}
\newacronym{VR}{VR}{Vitual Reality}
\newacronym{vnf}{VNFs}{Virtualized Network Functions}
\newacronym{vm}{VMs}{Virtual Machines}
\newacronym{vae}{VAE}{Variational Autoencoder}

\newacronym{wnd}{WND}{Wireless Networks Division}
\newacronym{wlan}{WLAN}{Wireless Local Area Network}
\newacronym{xApp}{xApp}{eXtended Application}
\newacronym{xr}{XR}{Extended Reality}

\newacronym{xapp}{xApp}{eXtended Application}

\usepackage{siunitx}
\begin{document}
\title{Benchmarking Unsupervised Jamming Detection in 5G Open RAN}
\author{Sneihil Gopal$^{\ddagger\dagger}$, Fernando J. Cintr\'{o}n$^{\star}$, and Richard A. Rouil$^{\star}$\\
$^{\ddagger}$PREP Associate, National Institute of Standards and Technology (NIST), USA\\
$^{\dagger}$Department of Physics, Georgetown University, USA\\
$^{\star}$National Institute of Standards and Technology (NIST), USA.\\
Emails: \{sneihil.gopal, fernando.cintron, richard.rouil\}@nist.gov}
\maketitle
\begin{abstract}
Radio Frequency (RF) jamming threatens fifth-generation (5G) networks. Open Radio Access Network (Open RAN) telemetry enables jamming detection without dedicated spectrum monitoring equipment, but labeled attack data are rarely available across deployments, motivating unsupervised detectors. Comparing such detectors across algorithmic families, however, remains difficult without ground-truth labels. This paper benchmarks statistical and reconstruction-based unsupervised jamming detectors using a symmetric evaluation interface that combines detection confidence and temporal interval coherence into a detector-independent composite score. We compare a Statistical Ensemble Method (SEM) detector against Vanilla, Long Short-Term Memory (LSTM), and Variational Autoencoder (VAE) baselines on Key Performance Indicator (KPI) telemetry from ten heterogeneous 5G environments with fixed and variable jamming bursts and dynamic jammer gains. Under leave-one-out (LOO) evaluation, SEM achieves a mean composite score of $\mathbf{0.906}$, compared with $\mathbf{0.688}$ for the best autoencoder (AE) per fold. Ground-truth validation yields a mean F1-score of $\mathbf{0.918}$ for SEM, compared with $\mathbf{0.521}$ for the best AE variant. The composite ranks SEM first in all environments, consistent with F1 rankings. However, in both within-trace and cross-trace evaluation, it fails for detectors that flag far fewer samples than the jamming prevalence, indicating that rankings are reliable only when detection rate matches the expected prevalence.
\end{abstract}
\section{Introduction}
\Gls{rf} jamming poses a persistent threat to \gls{5G} networks by disrupting legitimate transmissions at the physical layer without requiring protocol-level access~\cite{pirayesh2022jamming}. The disaggregated \gls{oran} architecture creates opportunities for interference detection through network telemetry, including \gls{ssb}-\gls{rsrp} statistics, the count of completed reference signal measurements, \gls{snr}, and \gls{rssi}~\cite{polese2023understanding}. These measurements provide observable signatures of jamming without additional spectrum monitoring equipment. However, their responses vary with jammer gain, burst duration, traffic conditions, and the number of connected \glspl{ue}. Furthermore, representative labeled attack data are rarely available across deployment environments, motivating detection methods that operate without labeled attack examples.

Two widely used families of detectors meet this requirement. Statistical ensembles flag deviations from adaptive baselines, whereas \glspl{ae} learn nominal behavior and flag high reconstruction error~\cite{ciocarlie2013detecting,ref_sakurada}. Their outputs, however, are not directly comparable, since ensembles express confidence through agreement among component detectors and \glspl{ae} through normalized reconstruction-error magnitude. Existing unsupervised evaluation methods assume a common anomaly-score representation or a single detector family \cite{ref_goix}, leaving cross-family comparison without ground truth largely unaddressed. To address this gap, we propose a symmetric evaluation interface that exposes only a binary detection mask and an associated confidence score from each detector. From these outputs, mean detection confidence and temporal interval coherence form a label-free composite score. However, a common interface does not imply equivalent confidence semantics or guarantee that higher composite scores correspond to better detection accuracy. For example, a detector that raises a few confident, contiguous alarms can score well while missing most jamming, and a detector that stays silent or flags almost everything is also unreliable. Composite rankings must therefore be checked against ground truth.

This paper performs that check on \gls{oran} telemetry from ten heterogeneous \gls{5G} environments with fixed and variable jamming bursts and dynamic jammer gains. It compares our previously proposed \gls{sem} detector~\cite{gopal2026oran} with Vanilla, \gls{lstm}, and \gls{vae} baselines under homogeneous (within-trace) and heterogeneous (cross-trace) evaluation, where the latter employs \gls{loo} validation to examine cross-environment generalization. In contrast to \cite{gopal2026oran}, where \gls{sem} was evaluated against supervised \gls{ml} and \Gls{DL} classifiers, in this work, we calibrate \gls{sem} without labels by maximizing an unsupervised composite score and compare it with detectors that also need no attack labels. Further, we use ground-truth annotations only to curate nominal \gls{ae} training and validation samples and to assess performance independently. They play no role in \gls{sem} calibration, \gls{ae} threshold selection, or composite score computation.

Our contributions are twofold. First, we propose a symmetric evaluation framework that compares statistical and reconstruction-based unsupervised jamming detectors while preserving each family's native calibration, and we apply it across ten heterogeneous \gls{5G} \gls{oran} environments. Second, we validate the unsupervised composite against independent ground-truth metrics. Under \gls{loo} evaluation, the composite ranks \gls{sem} (mean F1-score of $0.918$) above all \gls{ae} baselines (at most $0.521$) in every environment. Across the ten logs, \gls{sem}'s composite scores also correlate with its F1-scores. However, in both within-trace and cross-trace evaluation, composite score fails to rank detectors that flag far fewer samples than the true jamming prevalence, indicating that composite scores are reliable only when the detection rate is close to the expected prevalence.

The remainder of this paper is organized as follows. Section~\ref{sec:related} reviews related work. Section~\ref{sec:framework} presents the detector families and the evaluation framework. Section~\ref{sec:methodology} describes the experimental methodology. Section~\ref{sec:results} presents the results and 
Section~\ref{sec:conclusion} concludes the paper.
\section{Related Work}
\label{sec:related}
Statistical detection of cellular \gls{kpi} anomalies is well established. Robust z-score thresholding against a moving baseline is a standard approach~\cite{ref_iglewicz}, while ensemble methods fuse several such tests into one detector~\cite{ciocarlie2013detecting}. \gls{sem}~\cite{gopal2026oran} follows this line, making explicit the persistence filtering and interval merging such ensembles typically apply implicitly. The learning-based alternative, the \gls{ae}, trains on nominal data and flags high reconstruction error~\cite{ref_sakurada}, extended to time series via \gls{lstm}~\cite{ref_malhotra_lstm} and to a probabilistic form via \glspl{vae}~\cite{ref_kingma_vae}, with applications to \gls{iot} jamming detection~\cite{abdalzaher2021deep}.

\Gls{5G}-specific jamming detection has been investigated through classification of \gls{iq} samples using Bayesian methods \cite{jere2023bayesian} and spectrogram analysis using supervised and unsupervised models \cite{varotto2024detecting}. Closer to our setting, \gls{oran} testbeds detect jamming and signaling-storm anomalies from standard-interface telemetry \cite{kryszkiewicz2023open,bogucka2025open}, and our prior work compared \gls{sem} with supervised \gls{ml} and \gls{DL} classifiers across heterogeneous environments \cite{gopal2026oran}. These studies, however, calibrate or score detectors against labeled reference data and focus on detector design rather than on the reliability of label-free comparisons across detector families.

Unsupervised evaluation itself has received relatively little attention. Excess-Mass and Mass-Volume curves provide \gls{roc}- and \gls{pr}-consistent rankings in many cases without labels, though volume estimation restricts their applicability in higher dimensions \cite{ref_goix}. A composite of detection density and interval contiguity also anticipates one of our components \cite{ref_sipple}. Both approaches target a single detector or a shared anomaly-score representation, and neither establishes when label-free scores remain consistent with detection accuracy. This work addresses that gap. It checks label-free rankings of different detector families against the F1-score across heterogeneous \gls{ran} environments, and shows that these rankings are reliable only when a detector's detection rate matches the expected jamming prevalence.
\section{Proposed Framework}
\label{sec:framework}
This section presents the two detector families evaluated in this work, followed by the symmetric evaluation framework used for their comparison.

\subsection{SEM Detector}
\label{sec:sem}
\begin{figure}
    \centering
    \includegraphics[width=\columnwidth]
    {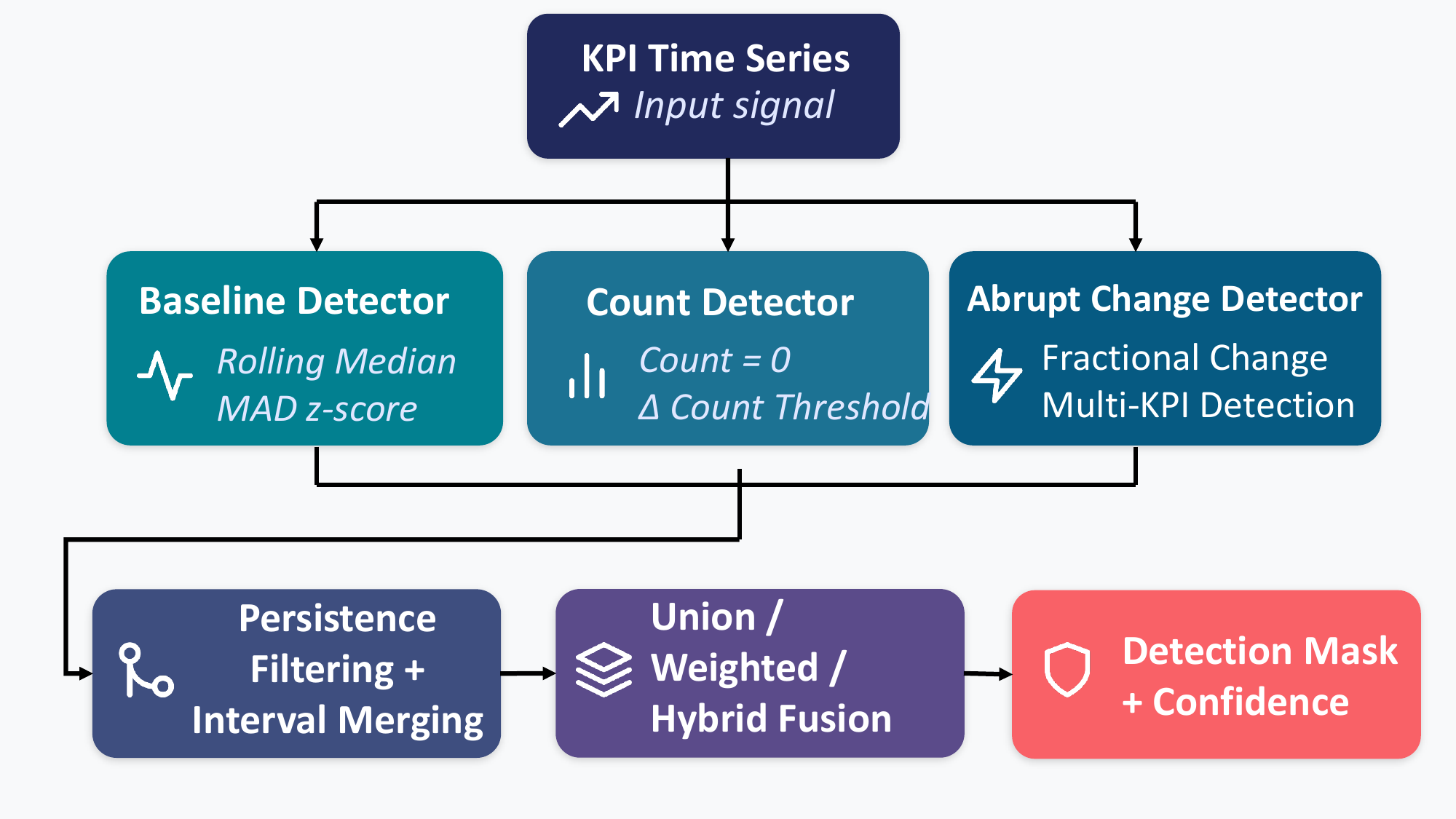}
    \caption{Overview of the \gls{sem} detection pipeline \cite{gopal2026oran}.}
    \label{fig:sem_pipeline}
\end{figure}
Figure~\ref{fig:sem_pipeline} summarizes the \gls{sem} pipeline \cite{gopal2026oran}. \gls{sem} operates on nine \glspl{kpi}: six \gls{ssb}-\gls{rsrp} statistics (minimum, first quartile, median, third quartile, maximum, and mean), the Count of completed \gls{ssb}-\gls{rsrp} measurements, \gls{snr}, and \gls{rssi}. It combines three complementary statistical sub-detectors: (i) the baseline detector $b$ flags robust z-score deviations from rolling median baselines, (ii) the count detector $c$ flags collapses and abrupt changes in the number of completed \gls{ssb}-\gls{rsrp} measurements, and (iii) the abrupt-change detector $a$ flags sudden \gls{kpi} transitions relative to each \gls{kpi}'s range. Each sub-detector $i \in \{b, c, a\}$ produces a binary mask $M_i = \{M_{i,t}\}$, where $M_{i,t} = 1$ flags sample $t$ as jamming.

\textbf{Post-processing:} Persistence filtering and interval merging suppress isolated flags and join nearby detections, rejecting runs shorter than $N$ samples and merging intervals separated by at most $G$ samples \cite{gopal2026oran}. The refined masks are denoted $M'_i$.

\textbf{Fusion:} Three fusion schemes, Union, Weighted, and Hybrid, combine the refined masks into a final detection mask $M$ and a per-sample confidence score
\begin{equation}
c_t = \frac{w_b M'_{b,t} + w_c M'_{c,t} + w_a M'_{a,t}}{w_b + w_c + w_a},
\end{equation}
where $(w_b, w_c, w_a)$ are the sub-detector weights. All three schemes report the same confidence score but differ in how the final mask is determined. Union declares a detection whenever any sub-detector fires, $M_t = M'_{b,t} \vee M'_{c,t} \vee M'_{a,t}$. Weighted thresholds the confidence directly, $M_t = \mathbbm{1}[c_t \geq \tau]$. Hybrid gives Count unconditional priority because $\text{Count}_t = 0$ is a strong signature of high-gain jamming under the considered testbed conditions, while requiring sufficient confidence before accepting detections from the remaining two sub-detectors:
\begin{equation}
M_t = M'_{c,t} \vee \left(M'_{b,t} \wedge (c_t \geq 0.5)\right) \vee \left(M'_{a,t} \wedge (c_t \geq 0.6)\right).
\end{equation}
Hybrid is used throughout the remainder of the paper. An ablation against the Union and Weighted rules is presented in Section~\ref{sec:ablation}.

\textbf{Unsupervised calibration:} The detector thresholds, post-processing lengths, fusion weights, and Weighted threshold $\tau$ defined in \cite{gopal2026oran} are selected independently for each trace through a grid search, with the rolling-window length held fixed. Instead of the supervised criterion used in \cite{gopal2026oran}, the search maximizes the \gls{sem}-native three-component composite defined in  Section~\ref{sec:eval_framework}, using only detector outputs and the observed \gls{kpi} time series.
 
\subsection{AE Baselines}
\label{sec:ae}
We consider three \gls{ae} baselines, namely Vanilla, \gls{lstm}, and \gls{vae}. Each learns a model of nominal \gls{kpi} behavior and flags samples whose reconstruction error is anomalously high. Unlike \gls{sem}, which builds its baseline directly from the test trace, the \glspl{ae} must first be trained on nominal data. This training data comes from the same trace in the homogeneous regime and from the other traces in the heterogeneous regime as described in Section~\ref{sec:methodology}.

\textbf{Architectures:} The three architectures span feedforward, temporal, and probabilistic reconstruction models trained with \gls{mse} loss. The Vanilla \gls{ae} uses a feedforward encoder-decoder with a $64$-dimensional latent space \cite{ref_sakurada}, the \gls{lstm} \gls{ae} encodes $10$-sample windows into a $32$-dimensional latent space to capture temporal dependencies \cite{ref_malhotra_lstm}, and the \gls{vae} learns a probabilistic $64$-dimensional latent representation \cite{ref_kingma_vae}. 

\textbf{Training:} The \glspl{ae} take $414$ temporal and statistical features engineered from the nine \glspl{kpi}, including rolling statistics, lags, differences, cross-\gls{kpi} ratios, and local spectral summaries, scaled to $[0, 1]$ on the training partition. They are trained for $30$ epochs with Adam (learning rate $10^{-3}$, batch size $32$) only on nominal samples curated with the jammer trigger signal, an idealized clean-data assumption that favors the \glspl{ae}. In contrast, \gls{sem} requires no training data.

\textbf{Threshold selection:} Each \gls{ae} computes a per-sample reconstruction error $e_t$, the discrepancy between the input at time $t$ and its reconstruction. The anomaly threshold is defined as $\theta=\bar{e}_{\mathrm{normal}}+k\,\sigma_{e_{\mathrm{normal}}}$, where $\bar{e}_{\mathrm{normal}}$ and $\sigma_{e_{\mathrm{normal}}}$ are the mean and standard deviation of the reconstruction error computed from the normal validation subset, and $k$ is a sensitivity multiplier. We sweep $k\in\{1.0,1.5,2.0,2.5,3.0\}$ and select the value maximizing the proposed unsupervised two-component composite on the complete, unlabeled validation partition, without using trigger labels. Under heterogeneous evaluation, this partition is drawn from the source traces, so the \glspl{ae} are never tuned on the target trace. 
Each test sample is classified as $M_t=\mathbbm{1}[e_t>\theta]$, producing a binary detection mask with the same semantics as in Section~\ref{sec:sem}. A confidence score $c_t=(e_t-e_{\min})/(e_{\max}-e_{\min})$ is then computed and clipped to $[0,1]$, where $e_{\min}$ and $e_{\max}$ denote the minimum and maximum reconstruction errors over the test partition. 

\subsection{Evaluation Framework}
\label{sec:eval_framework}
This subsection presents the symmetric evaluation framework used to compare both detector families while preserving their native calibration procedures.

\textbf{Design properties:} A metric supporting architecture-independent comparison of unsupervised detectors should satisfy four properties. Only metrics satisfying all four qualify as composite components.
\begin{itemize}[leftmargin=*]
    \item \textbf{P1. Architecture-independent computability.} Depends only on the detector output (mask and confidence) and the observed \gls{kpi} series, not on internal representations like sub-detector masks or latent-space reconstruction error.
    \item \textbf{P2. Boundedness.} Lies in a fixed range, here $[0, 1]$, so detectors with different confidence scales can be aggregated.
    \item \textbf{P3. Detection-conditioned monotonicity.} Higher anomaly confidence and greater temporal coherence should not decrease the score.
    \item \textbf{P4. Failure-mode visibility.} Detectors producing no detections (silence) are distinguished from detectors producing poor detections, since these require different operational responses.
\end{itemize}
Two component scores satisfy all four properties. The mean confidence $c_{\text{conf}}$ is the average detector confidence over all detected samples. The interval coherence $c_{\text{coh}}$ combines interval length and fragmentation, rewarding long contiguous detections. P1 and P2 ensure that the confidence measures are computable and share a common range, not that they have identical semantics. \gls{sem}'s confidence reflects sub-detector agreement, whereas \gls{ae} confidence reflects normalized reconstruction-error magnitude. Accordingly, $c_{\text{conf}}$ compares the strength of each detector's exported confidence signal rather than architecture-independent anomaly evidence. Unlike conventional formulations that assign zero to empty detection sets, the proposed framework reports the composite as undefined, explicitly identifying silence (P4). Because P3 conditions both components on detected samples, neither measures how much of the jamming activity is covered. This coverage gap is addressed by the detection-rate diagnostic below and examined empirically in Section~\ref{sec:ranking_consistency}.

Several intuitive alternatives fail at least one property. Output entropy is not detection-conditioned (P3 and P4), coverage depends on \gls{kpi}-specific normalization (P2), and detection rate is bounded but is not monotonically related to detector quality because both silence and saturation can correspond to poor operational behavior. Training stability based on repeated model fitting is not naturally applicable to deterministic statistical detectors such as \gls{sem} (P1), while cross-detector consensus is undefined for monolithic detectors like the \glspl{ae} (P1).

\textbf{Component scores and composites:} For \gls{sem}, $c_{\text{conf}}$ is the weighted vote ratio defined in Section~\ref{sec:sem}, and for each \gls{ae} it is the normalized reconstruction error defined in Section~\ref{sec:ae}. Given $N_{\text{det}}$ detected samples grouped into $n_{\text{int}}$ contiguous intervals of mean length $\bar{L} = N_{\text{det}}/n_{\text{int}}$,
\begin{equation}
c_{\text{coh}} = 0.6\,\min\!\left(\frac{\bar{L}}{5}, 1\right) + 0.4\,\max\!\left(0, 1 - \frac{2}{\bar{L}}\right),
\end{equation}
where the weights and the five-sample saturation length are set heuristically. The head-to-head composite is
\begin{equation}
C_{\text{2-comp}} = \frac{c_{\text{conf}} + c_{\text{coh}}}{2},
\label{eq:c2comp}
\end{equation}
and it is reported as undefined when a detector produces no detections. The \gls{sem}-native composite
\begin{equation}
C_{\text{3-comp}} = \frac{c_{\text{conf}} + c_{\text{coh}} + c_{\text{kpi}}}{3}
\label{eq:c3comp}
\end{equation}
adds the cross-\gls{kpi} consistency $c_{\text{kpi}}$, the mean fraction of the per-\gls{kpi} abrupt-change masks that flag each detected sample. Because $c_{\text{kpi}}$ requires \gls{sem}'s internal per-\gls{kpi} masks, it violates P1 and is used only for \gls{sem} hyperparameter selection.

\textbf{Standalone diagnostics:} Two diagnostics are reported alongside the composite but intentionally excluded from it. The detection rate $N_{\text{det}}/N$, where $N$ is the number of evaluated samples, is compared against an operator prior on jamming prevalence to identify silence, under-detection, and saturation. The interval count $n_{\text{int}}$ characterizes alarm fragmentation. Neither satisfies P3, but Section~\ref{sec:ranking_consistency} shows that the detection rate is required to determine when composite rankings can be trusted.
\section{Experimental Methodology}
\label{sec:methodology}
This section describes the telemetry and threat model, the ten-trace dataset and its cell-level aggregation, and the two cross-validation regimes used to evaluate both detector families. It also defines the ground-truth metrics used to validate the unsupervised composite independently of detector calibration.

\textbf{Telemetry and KPIs:} We consider a \gls{5G} \gls{oran} deployment operating in Band n78 ($3.3$ to $3.8$~GHz) with one or more \glspl{ue} under downlink traffic. The \gls{gnb} exposes the nine per-measurement-interval \glspl{kpi} of Section~\ref{sec:sem} through standard \gls{oran} interfaces. \gls{sem} operates on these \glspl{kpi} directly, whereas the \glspl{ae} receive the $414$ features engineered from them (Section~\ref{sec:ae}).

\textbf{Threat model:} The threat is an external \gls{rf} jammer transmitting into the cell's operating band and cycled on and off, with fixed or dynamically varying gain. Jamming disrupts reception of the \glspl{ssb} from which the \gls{rsrp} measurements are derived, so that under high-gain jamming the receiver may fail to complete the measurement procedure and Count collapses toward zero. The direction of the \gls{ssb}-\gls{rsrp} shift is testbed-dependent, since the receiver may lose the legitimate signal (\gls{rsrp} decreases) or lock onto the jammer (\gls{rsrp} increases). \gls{snr} and \gls{rssi} exhibit similar variability.

\begin{table}[t]
\centering
\vspace*{0.08in}
{\small
\begin{tabular}{p{0.35cm}p{0.5cm}p{1cm}p{1cm}p{0.75cm}p{1.35cm}p{1cm}}
\hline
\textbf{Log} & \textbf{UEs} & \textbf{Duration} & \textbf{ON (s)} & \textbf{Gain (dB)} & \textbf{Samples} & \textbf{Jam (\%)} \\
\hline
$1$  & $3$ & $68$ min  & $60$       & $48$       & $\num{12185}$ & $56.3$ \\
$2$  & $2$ & $62$ min  & $60$       & $48$       & $\num{7467}$  & $56.4$ \\
$3$  & $2$ & $43$ min  & $60$       & $48$       & $\num{5131}$  & $57.7$ \\
$4$  & $1$ & $47$ min  & $60$       & $48$       & $\num{2810}$  & $59.1$ \\
$5$  & $1$ & $79$ min  & $30$       & $48$       & $\num{4749}$  & $40.8$ \\
$6$  & $1$ & $15$ min  & $60$       & $45$       & $\num{925}$   & $58.8$ \\
$7$  & $1$ & $3.1$ h   & $[10, 30]$ & $45$       & $\num{10980}$ & $20.5$ \\
$8$  & $1$ & $18.1$ h  & $[10, 30]$ & $45$       & $\num{65074}$ & $22.7$ \\
$9$  & $3$ & $2.0$ h   & $[5, 14]$  & $[48, 60]$ & $\num{17465}$ & $26.0$ \\
$10$ & $3$ & $6.3$ h   & $[5, 14]$  & $[48, 60]$ & $\num{59028}$ & $19.9$ \\
\hline
\end{tabular}
\caption{Dataset summary. Samples are per-\gls{ue} reports, and Jam is the fraction of labeled reports under jamming.}
\label{tab:dataset}}
\end{table}

\textbf{Dataset:} The dataset comprises ten \gls{kpi} traces collected from a \gls{5G} \gls{oran} testbed and summarized in Table~\ref{tab:dataset}, spanning $34.7$ hours of heterogeneous operating conditions. Logs $1$ to $8$ use constant gains of $45$ or $48$~dB with fixed or variable burst durations, while Logs $9$ and $10$ use shorter irregular bursts with gains varying between $48$ and $60$~dB. The traces also vary the number of attached \glspl{ue} and the recording duration. For multi-\gls{ue} configurations, the $\num{185814}$ per-\gls{ue} reports are aggregated into $\num{123040}$ cell-level observations by timestamp. Count and minimum \gls{rsrp} use the minimum across \glspl{ue}, maximum \gls{rsrp} uses the maximum, and the remaining channels use the mean. The ground-truth state is combined using logical OR across \glspl{ue}. 
Ground-truth labels derived from the jammer trigger are used only to curate the \gls{ae} nominal training and validation subsets (Section~\ref{sec:ae}) and for independent performance assessment. They are never used for \gls{sem} calibration, selection of the \gls{ae} multiplier $k$, detector inference, or composite computation.

\textbf{\gls{cv} protocol:} Two regimes assess detector behavior within and across environments. In the \emph{homogeneous regime} (single-file \gls{cv}), each trace is evaluated independently. \gls{sem} operates retrospectively on the full trace, while the \glspl{ae} use a $70/15/15$ temporal train/validation/test split with nominal-only training and are evaluated on the test partition. 
\gls{sem} predictions are restricted to the same test observations, and Log~$10$ is excluded because its test partition contains no jamming. In the \emph{heterogeneous regime} (multi-file \gls{loo} \gls{cv}), each of the ten traces is held out once. The \glspl{ae} train on nominal samples from the remaining nine traces, validate on their chronologically last $15\,\%$, and are applied to the held-out trace without retraining, while \gls{sem} operates retrospectively on the held-out trace. Because \gls{sem} is calibrated directly on each evaluation trace, this regime tests cross-environment transfer for the \glspl{ae} but not source-to-target parameter transfer for \gls{sem}.

\textbf{Independent performance validation:} All detection results are computed on the cell-level observations. Detector outputs are evaluated retrospectively against ground truth using precision, recall, F1-score, F2-score, event recall, false-alarm intervals per $\num{1000}$ evaluated samples, and onset delay. 
A jamming event is a contiguous labeled interval, detected if any alarm interval overlaps it. A false-alarm interval overlaps no event, and onset delay is the samples from event start to first alarm. Paired Wilcoxon signed-rank tests across folds assess whether F1-score differences between detectors are significant. Agreement between $C_{\text{2-comp}}$ and F1-score rankings is assessed per fold and through Spearman and Kendall correlations pooled over all scored detector-fold pairs. Finally, the ratio $r$ of detection rate to jamming prevalence identifies when composite rankings are reliable.
\section{Results and Discussion}
\label{sec:results}
This section compares the detectors under the composite and ground truth in both regimes, examines when composite rankings agree with ground truth and the prevalence condition for their reliability, and closes with the fusion ablation, limitations, and \gls{oran} deployment implications.

\subsection{Composite and Ground-Truth Performance}
\label{sec:unsup_eval}
\begin{figure}
    \centering
    \includegraphics[width=\columnwidth]{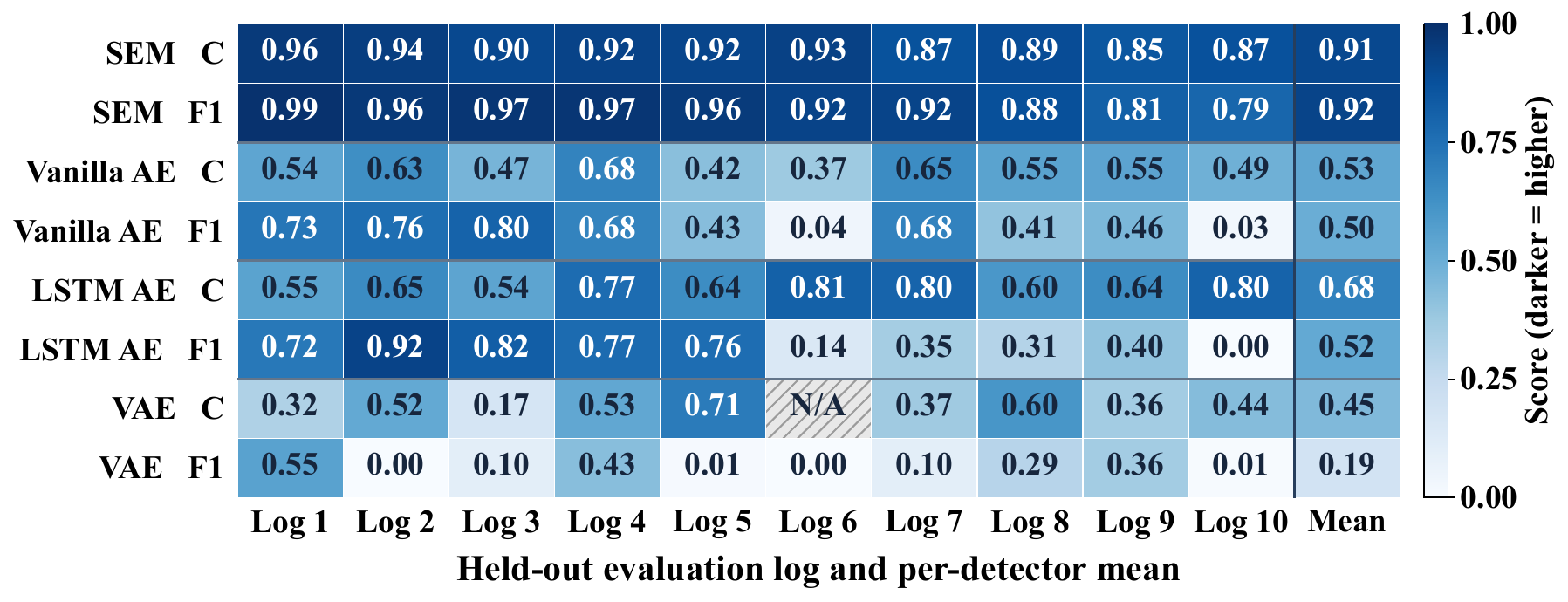}
    \caption{Per-log composite ($C$) and F1 under \gls{loo} evaluation.}
    \label{fig:heatmap}
\end{figure}
\textbf{Heterogeneous evaluation:} In Figure~\ref{fig:heatmap}, \gls{sem}'s mean $C_{\text{2-comp}}$ is $0.906$ versus $0.688$ for the best \gls{ae} per fold, and it ranks first in all $10$ environments, including the dynamic-gain Logs $9$ and $10$. Ground truth confirms this lead: \gls{sem} reaches a mean F1-score of $0.918$ (F2 of $0.934$) versus $0.521$, $0.503$, and $0.185$ for the \gls{lstm} \gls{ae}, Vanilla \gls{ae}, and \gls{vae}, beating every \gls{ae} in all ten folds (two-sided Wilcoxon signed-rank $p = 0.002$ against each). It also catches almost every jamming event (event recall of $0.978$) with $0.302$ false-alarm intervals per $\num{1000}$ samples, whereas the \glspl{ae} miss entire events, with event recall of $0.83$, $0.72$, and $0.47$ for the Vanilla \gls{ae}, \gls{lstm} \gls{ae}, and \gls{vae}. \gls{sem}'s F1-score exceeds $0.9$ on Logs $1$ to $7$ and drops to $0.876$, $0.814$, and $0.793$ on Logs $8$ to $10$. On Logs $9$ and $10$, the short $5$ to $14\,$s bursts break its detections into $188$ and $304$ intervals and lower its mean confidence to $0.77$ and $0.81$, making short irregular bursts its hardest case. The drop comes from precision ($0.74$ to $0.78$), while recall stays high ($0.85$ to $1.00$), so \gls{sem} tends to raise extra alarms rather than miss jamming, the safer failure mode for mission-critical networks. \gls{sem} also alarms sooner ($0.8$ versus $3.8$ to $7.0$ samples after onset), and its detection rate stays within $0.91$ to $1.28$ times the jamming prevalence, so it neither under-detects nor over-detects on any log. The \gls{vae} is silent on Log~$6$, so its composite is undefined.

\textbf{Homogeneous evaluation:} \gls{sem}'s composite is again $0.906$, as its calibration is identical in both regimes. The best-\gls{ae} composite rises to $0.798$ over the $9$ folds with a detecting \gls{ae} (no \gls{ae} detects on Log~$6$), suggesting that the \glspl{ae} improve when trained on the test environment. Ground truth shows otherwise. On the matched test partitions of the nine traces with jamming, \gls{sem} reaches a mean F1 of $0.885$ versus $0.501$, $0.505$, and $0.505$ for the Vanilla \gls{ae}, \gls{lstm} \gls{ae}, and \gls{vae}, and again beats every \gls{ae} on every trace ($p = 0.004$), so a higher \gls{ae} composite does not mean better detection. The reason is under-detection. With one training trace and a short test partition ($139$ to $\num{9762}$ samples), the composite favors high thresholds and selects the largest multiplier, $k = 3.0$, in $20$ of the $30$ cases. The \glspl{ae} then flag only $24\,\%$ to $28\,\%$ of samples, although jamming occupies $43\,\%$ on average, and detect only $45\,\%$ to $49\,\%$ of it. No \gls{ae} that flags less than $40\,\%$ of the jamming prevalence ($r < 0.4$) exceeds an F1 of $0.53$, the failure mode analyzed next.

\subsection{Consistency of Unsupervised Rankings}
\label{sec:ranking_consistency}
\textbf{Across families and environments:} In all $10$ \gls{loo} folds, the composite ranks \gls{sem} above every \gls{ae}, as does the F1-score. Pooled over all $39$ scored detector-fold pairs (four detectors on ten folds, excluding the silent \gls{vae} on Log~$6$), higher composite scores go with higher F1-scores (Spearman $0.63$, $p < 0.001$, Kendall $\tau = 0.45$). This alone is weak evidence, since it mainly reflects the large gap between \gls{sem} and the \glspl{ae}. A stricter test looks at \gls{sem} alone: on logs where its composite is lower, its F1-score is also lower (Spearman $0.76$, $p = 0.01$), so the composite reflects how difficult each environment is.

\textbf{Within the \gls{ae} family:} Here the composite is unreliable. The highest-scoring \gls{ae} is the most accurate in only $4$ of $10$ folds, and over the $29$ scored \gls{ae} results, composite and F1-score are nearly uncorrelated (Spearman $0.15$, $p = 0.45$). The worst case is Log~$10$, where the \gls{lstm} \gls{ae} has the highest \gls{ae} composite ($0.80$) yet detects only $5$ of $\num{3291}$ jamming samples (F1 of $0.003$). The \gls{lstm} \gls{ae} on Log~$6$ ($0.81$, F1 of $0.14$) and the \gls{vae} on Log~$5$ ($0.71$, F1 of $0.01$) behave similarly. These \glspl{ae} raise a few confident, contiguous alarms, which $c_{\text{conf}}$ and $c_{\text{coh}}$ reward, while missing nearly all jamming. The cause is threshold transfer: thresholds learned on other traces can be far too low or too high for the target, so the Vanilla and \gls{lstm} \glspl{ae} flag $94\%$ and $99.5\%$ of Log~$1$ samples but fewer than $1\%$ of Log~$10$ samples. The composite penalizes over-detection, because flagging nearly every sample includes many low-error samples and lowers $c_{\text{conf}}$ (composites of $0.54$ to $0.55$), but not under-detection, because the few flagged samples are the highest-error ones and keep $c_{\text{conf}}$ high.

\begin{table}[t]
\vspace*{0.08in}
\centering
{\small
\begin{tabular}{p{3cm}p{1cm}p{1cm}p{1cm}p{0.75cm}}
\toprule
Group & Results & Median $C_{\text{2-comp}}$ & Median F1 & Max F1 \\
\midrule
\gls{ae}, $r < 0.3$ & $12$ & $0.51$ & $0.07$ & $0.43$ \\
\gls{ae}, $r \geq 0.5$ & $17$ & $0.60$ & $0.68$ & $0.92$ \\
\gls{sem}, $r \in [0.91, 1.28]$ & $10$ & $0.91$ & $0.94$ & $0.99$ \\
\bottomrule
\end{tabular}
\caption{Composite and F1 grouped by the ratio $r$ of detection rate to jamming prevalence under \gls{loo} evaluation.}
\label{tab:prevalence_regime}}
\end{table}

\textbf{Reliability regime:} Table~\ref{tab:prevalence_regime} groups results by $r$ to show when the composite can be trusted. All failures above fall in the \gls{ae} group with $r < 0.3$, which flags less than $30\%$ of the jamming present and has a median F1-score of only $0.07$. Yet its median composite ($0.51$) is close to that of the \gls{ae} group with $r \geq 0.5$ ($0.60$), whose median F1-score is $0.68$. The composite therefore cannot tell under-detecting detectors from adequate ones. Knowing the prevalence would largely fix this: as a hindsight bound, multiplying $C_{\text{2-comp}}$ by $\min(r, 1/r)$ raises best-\gls{ae} agreement from $4$ to $8$ of $10$ folds and the pooled \gls{ae} Spearman correlation from $0.15$ to $0.80$. This uses ground truth and is not deployable, but it shows that even a coarse operator estimate of prevalence is the most useful complement to the composite. The composite thus reliably separates detector families, but choosing among similar detectors also requires checking the detection rate.

\subsection{Fusion Ablation and Discussion}
\label{sec:ablation}

\textbf{Fusion ablation:} Hybrid is the best rule, with a mean F1-score of $0.918$ versus $0.854$ for Weighted and $0.670$ for Union, and $0.302$ false-alarm intervals per $\num{1000}$ samples versus $31.1$ and $22.4$. Union fails because it accepts every brief single-\gls{kpi} shift flagged by the abrupt-change sub-detector, flagging every labeled sample of Log~$6$ and about $70\,\%$ of Logs $7$ and $8$, although jamming occupies only about $20\,\%$ of those logs. Hybrid avoids this by always accepting Count-based detections while requiring confidence for the others, keeping recall high without the false alarms. 

\textbf{Limitations:} \gls{sem} is calibrated on each evaluation trace, whereas the \glspl{ae} transfer from other environments, so the two families operate under different assumptions. Confidence scores share a range but not semantics, and $C_{\text{2-comp}}$ penalizes neither under-detection nor false alarms. The composite-selected \gls{ae} thresholds often lie at the sweep boundary, with $k = 3.0$ for the \gls{vae} in $9$ of $10$ \gls{loo} folds and $k = 1.0$ for the Vanilla \gls{ae} in $6$, so a wider sweep may shift individual \gls{ae} results. Finally, all traces come from a single testbed, and causal operation and live evaluation remain future work.
\begin{figure}[t]
\centering
\includegraphics[width=\columnwidth]{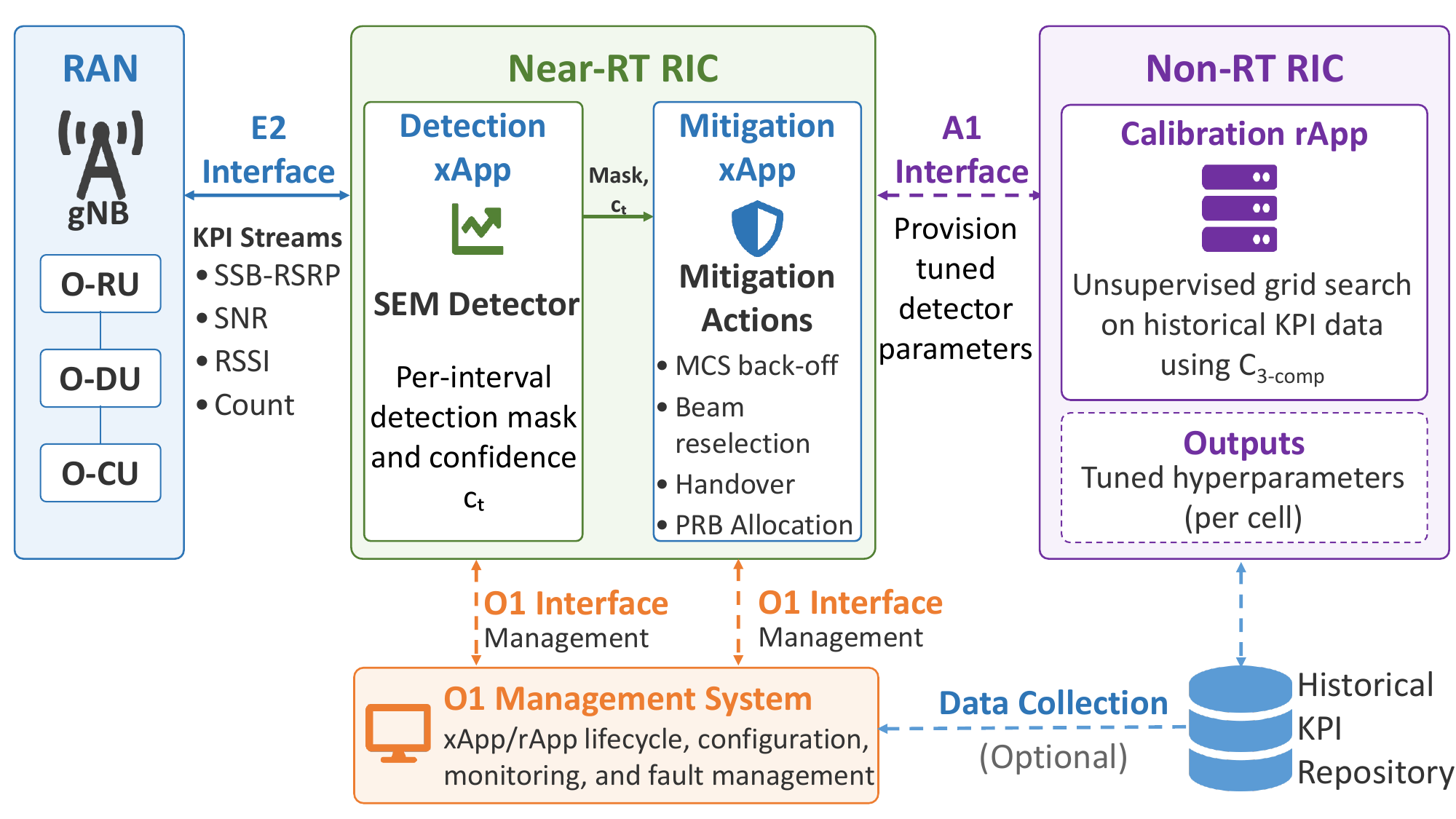}
\caption{Proposed \gls{sem} deployment in the \gls{oran} architecture.}
\label{fig:deployment}
\end{figure}
\subsection{\gls{oran} Deployment Implications}
\label{sec:deployment}
Figure~\ref{fig:deployment} illustrates how \gls{sem} can be deployed within the \gls{oran} architecture. \gls{sem} can run as a detection \gls{xapp} on the Near-\gls{rt} \gls{ric}, consuming \gls{kpi} streams over E2. Because its statistics are centered, it would decide over a buffered window, and this buffering latency is not included in the onset delay of Section~\ref{sec:unsup_eval}. A calibration \gls{rapp} on the Non-\gls{rt} \gls{ric} can run the unsupervised grid search on historical telemetry and provision per-cell parameters over A1, so no stage would require labeled attack data. The confidence score can support graded mitigation, from \gls{mcs} adaptation for weak detections to beam reselection, handover, or \gls{prb} reallocation for sustained strong ones. The same \gls{rapp} can also compare candidate detectors with $C_{\text{2-comp}}$, accepting a ranking only when each detection rate is consistent with the expected prevalence.

\section{Conclusion}
\label{sec:conclusion}
This paper benchmarked statistical and reconstruction-based unsupervised jamming detectors on ten \gls{oran} telemetry logs using a symmetric composite of detection confidence and interval coherence. Under \gls{loo} evaluation, \gls{sem} achieved a mean composite of $0.906$ and F1-score of $0.918$, compared with $0.688$ and $0.521$ for the best \gls{ae}, and ranked first in all ten environments under both measures. 
The composite tracked how \gls{sem} performance varied across environments. However, in both regimes it ranked \glspl{ae} incorrectly when they flagged far fewer samples than the true jamming prevalence, so composite rankings are reliable only when detection rate matches the expected prevalence. Future work will evaluate additional testbeds, jammer behaviors, and detector families, and will explore prevalence-aware scoring and live evaluation within the \gls{oran} architecture.
\section*{Acknowledgment}
This work is partially funded by the Department of Homeland Security’s Science and Technology Directorate (S\&T). U.S. Government work, not subject to U.S. Copyright.
\begin{spacing}{}
\bibliographystyle{IEEEtran}
\bibliography{references}
\end{spacing}
\end{document}